\documentclass[letterpaper]{article} 
\usepackage{aaai23}  
\usepackage{times}  
\usepackage{helvet}  
\usepackage{courier}  
\usepackage[hyphens]{url}  
\usepackage{graphicx} 
\usepackage{natbib}  
\usepackage{caption} 
\usepackage{algorithm}
\usepackage{algorithmic}

\usepackage{newfloat}
\usepackage{listings}
\DeclareCaptionStyle{ruled}{labelfont=normalfont,labelsep=colon,strut=off} 
\floatstyle{ruled}
\newfloat{listing}{tb}{lst}{}
\floatname{listing}{Listing}
\title{Build-a-Bot: Teaching Conversational AI Using a Transformer-Based Intent Recognition and Question Answering Architecture}

\author {
    Kate Pearce\textsuperscript{\rm 1},
    Sharifa Alghowinem \textsuperscript{\rm 2},
    Cynthia Breazeal \textsuperscript{\rm 2}
}
\affiliations {
    \textsuperscript{\rm 1} Massachusetts Institute of Technology\\
    \textsuperscript{\rm 2} MIT Media Lab\\
    k8pearce@mit.edu, sharifah@media.mit.edu, cynthiab@media.mit.edu
}

\begin{document}

\maketitle

\begin{abstract}
As artificial intelligence (AI) becomes a prominent part of modern life, AI literacy is becoming important for all citizens, not just those in technology careers. Previous research in AI education materials has largely focused on the introduction of terminology as well as AI use cases and ethics, but few allow students to learn by creating their own machine learning models. Therefore, there is a need for enriching AI educational tools with more  adaptable and flexible platforms for interested educators with any level of technical experience to utilize within their teaching material. As such, we propose the development of an open-source tool (Build-a-Bot) for students and teachers to not only create their own transformer-based chatbots based on their own course material, but also learn the fundamentals of AI through the model creation process. The primary concern of this paper is the creation of an interface for students to learn the principles of artificial intelligence by using a natural language pipeline to train a customized model to answer questions based on their own school curriculums. The model uses contexts given by their instructor, such as chapters of a textbook, to answer questions and is deployed on an interactive chatbot/voice agent. The pipeline teaches students data collection, data augmentation, intent recognition, and question answering by having them work through each of these processes while creating their AI agent, diverging from previous chatbot work where students and teachers use the bots as black-boxes with no abilities for customization or the bots lack AI capabilities, with the majority of dialogue scripts being rule-based. In addition, our tool is designed to make each step of this pipeline intuitive for students at a middle-school level. Further work primarily lies in providing our tool to schools and seeking student and teacher evaluations.

\end{abstract}

\section{Introduction}

Within the past few years, quality STEM education has become an ever-pressing issue, especially in regards to increasing the representation of racial and gender minorities within STEM fields \cite{stem}. Artificial intelligence (AI) education, in particular, has become necessary not only for prospective computer scientists and engineers, but for everyone: as AI technologies become facets of daily life, AI literacy becomes a crucial skill for the next generation.

Much of the current work in AI education centers around curriculum design and classroom implementation -- for example, MIT's AI Ethics for Middle School curriculum focuses on showing students examples of AI that most of them regularly use, developing their programmatic thinking skills, and raising their awareness of ethical issues concerning AI like algorithmic bias \cite{ethics}. Current literature is primarily focused on lesson and activity design to introduce students to AI terminology and the computational mindset \cite{competencies} as well as broad perspective and ethical issues related to AI \cite{aiEdSurvey}.

While previous research has focused on curriculum development, this paper proposes a tool for AI education that would allow students to explore the supervised learning process by creating their own chatbots using a transformer-based language pipeline. Our proposed tool explores the following:
\begin{itemize}
  \item The use of multiple transformer models to create a sustainable and scalable pipeline for human-AI educational interaction, utilizing the following:
  \begin{itemize}
      \item Data augmentation methods for question-based dialogue.
      \item An intent recognition model to classify student questions by topic.
      \item Extractive and generative question answering models for answering student questions given a context about the topic the question is labeled as.
  \end{itemize}
  \item Providing an open-source tool to facilitate constructivist learning of AI in the classroom, which enables the following:
  \begin{itemize}
      \item AI education by instructors of any level of technical expertise.
      \item Student development of problem-solving skills by having them understand and engage with a complex system.
  \end{itemize}
\end{itemize}

\section{Educational Background}

\subsection{AI Education}

 As artificial intelligence (AI) becomes increasingly powerful, and its applications grow to span almost every field, the development of an AI education curriculum grows increasingly relevant. Previous works \cite{competencies} have sought to create an educational framework for AI and have established three AI competencies for primary and secondary students: knowledge, skill, and attitude. While AI knowledge is described as knowing the definition and types of AI, AI skill revolves around knowing how to use AI tools and AI attitude around describing the impact of and collaborating with AI. The goal of their framework is to develop student's AI literacy: even students who don't pursue careers as technologists should have a basic understanding of how AI works and how it is engaged with.

\begin{figure}[t]
    \centering
    \includegraphics[width=\linewidth]{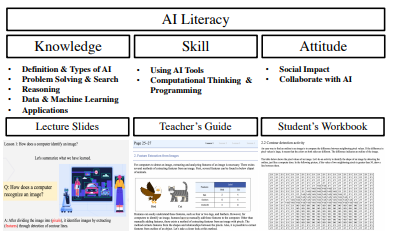}
    \caption{Sample elementary school AI curriculum based on core competencies of knowledge, skill, and attitude \cite{competencies}.}
\end{figure}

Existing curriculums to build AI literacy at the elementary and middle school level largely focus on programmatic thinking and applications of AI: Google's Teachable Machine is a popular tool to demonstrate the process of learning from a dataset, and students may be more directly instructed on examples of AI seen in daily life, like YouTube's recommendation system \cite{middleCurriculum}. While these methods do well in enhancing student awareness of AI, gaps exist between what is being taught by educators and what is desired by AI practitioners; for example, while educators often focus on puzzles and games to illustrate AI, practitioners largely cite understanding and engineering systems as a necessary learning objective \cite{aiEdSurvey}.

\begin{figure}[t]
    \centering
    \includegraphics[width=\linewidth]{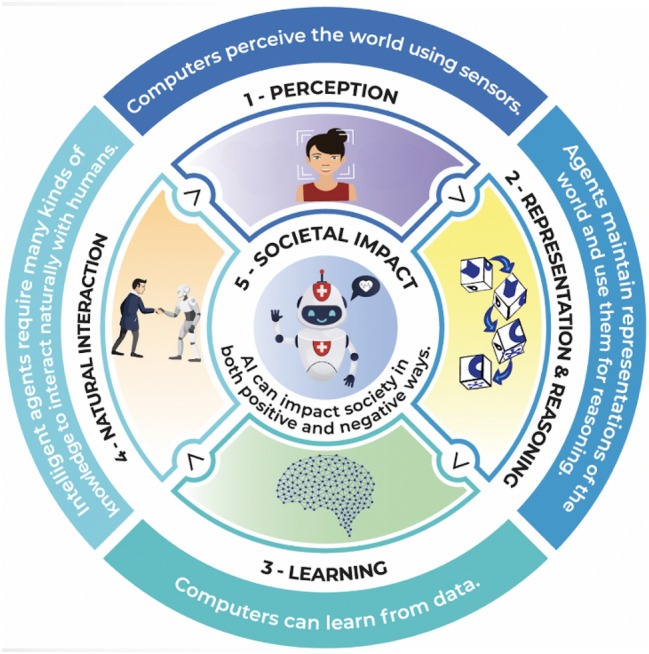}
    \caption{The AI4K12 initiative's Big Ideas of AI. Build-a-Bot allows students to "tinker" with AI models in order to understand these concepts \cite{bigideasgraphic}.}
    \label{fig:ideas}
\end{figure}
 
The AI4K12 initiative has developed AI curriculum based on "big ideas" that every K12 students should know about AI based on consultations with both AI experts and K-12 educators. These themes are broad guidelines for AI education and include "Perception", "Representation and Reasoning", "Learning", "Natural Interaction", and "Societal Impact". Build-a-Bot primarily serves an an educational tool for "Learning" and "Natural Interaction", detailed in Figure \ref{fig:ideas}, as students learn the process of how models can be trained to answer questions and develop a chatbot that they can directly interact with using those models \cite{bigideas}.
\subsection{Constructivist Educational Model}
Constructivism refers to a theory of learning in which students construct unique conceptual understandings for themselves, rather than simply being "handed" these understandings by a teacher \cite{piaget}. Constructivists typically view traditional lecture modes of instruction as promoting rigid and simplified knowledge structures that hinder future learning and preventing students from engaging with the full nuance of topics. Moreover, social constructivism, a variant of constructivism, is a framework that has risen in popularity in recent educational history in which students construct conceptual understandings through interactions with each other, including discussion and collaborative problem-solving \cite{flippedActive}.


Much recent work in educational methodology has centered around approaches to augment the constructivist model in the classroom. Student-centered learning environments, for example, tailor instruction to each student's individual needs and perspectives; students are to take full responsibility for their learning and are engaged by connecting the curriculum to their interests in the world at large \cite{studentCenter}. While active learning has been rather loosely defined in current literature, it serves to complement the responsibility aspect of student-centered approaches by having students learn by solving problems and engaging in other creative activities in order to build knowledge, as opposed to passive lecture-based instruction \cite{threeTheories}. There is a substantial body of literature proving the efficacy of active learning, including increased content retention as well as improvements in student self-esteem and engagement in learning \cite{prince}. Transformational teaching, meanwhile, is an approach to implementing the social constructivist model in which instructors promote dynamic relationships between students and themselves in order to help students master course concepts while also improving their attitudes towards learning itself \cite{transform}.

Build-a-Bot implements these methodologies in its design; students learn by studying and modifying a complex language system, rather than being directly instructed on AI principles. Students are able to make chatbots on any topic of interest and develop a new understanding of how AI is used and why it is useful throughout the chatbot's creation.

\section{Technical Background}

\subsection{Conversational AI}

The purpose of conversational AI is to develop an agent that can engage in human-like conversation that is both informative and controllable, meaning that answers should be grounded in external knowledge and generated in a specified style \cite{convoAI}. Models should be able to both understand questions and generate appropriate responses, as well as understand conversational history in order to engage in multi-turn dialogue effectively \cite{convoAI2}.

\begin{figure*}[t]
    \centering
    \includegraphics[width=\linewidth]{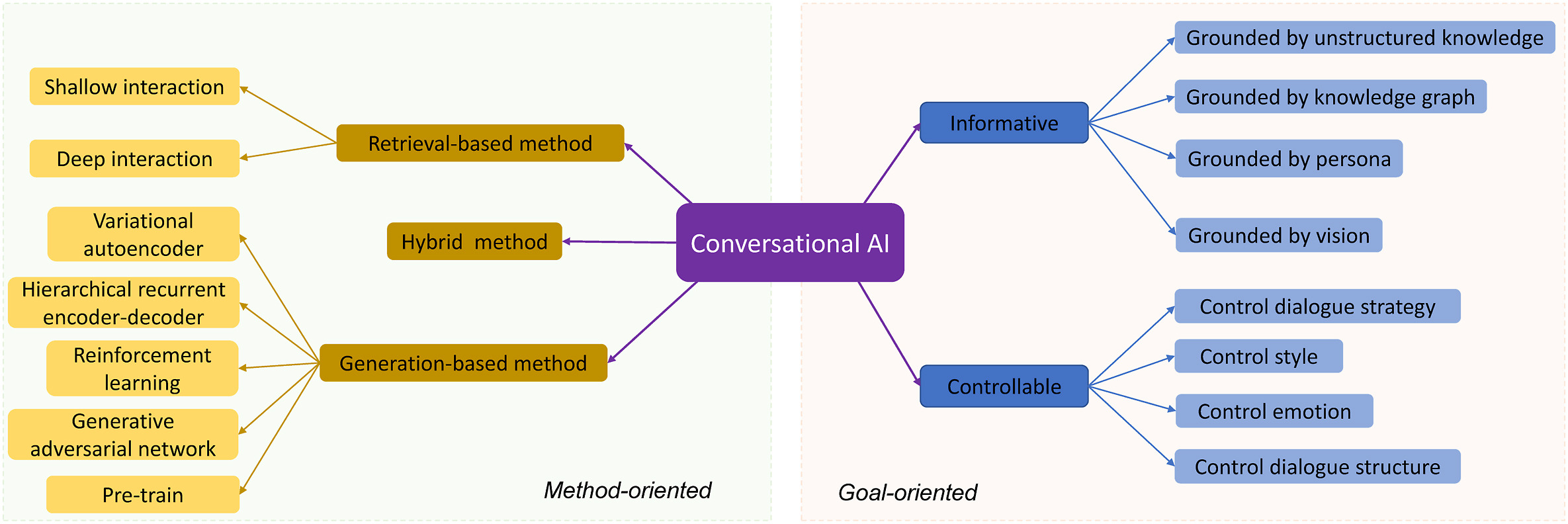}
    \caption{Tree diagram depicting the types and goals of conversational AI; retrieval-based methods evaluate the relevancy of pre-written responses, while generative approaches generate new responses. \cite{convoAI}.}
\end{figure*}

\subsection{Transformers}
Though recurrent neural networks and long short-term memory have been established as language modeling approaches, these models factor computation along symbol positions of input and output sequences, which precludes parallelization among training samples. Bidirectional versions of these models, which encode sentences from both start to end and end to start, have been used to mitigate this problem, though this doesn't substantially increase performance for long dependencies \cite{bidirectionality}. Attention mechanisms, which allow modeling of dependencies without consideration of their distance within input and output sequences, have been introduced to combat this problem; the transformer is the first model to rely entirely on the use of attention mechanisms without being used alongside a recurrent neural network \cite{attention}. 

\begin{figure}[!h]
    \centering
    \includegraphics[width=\linewidth]{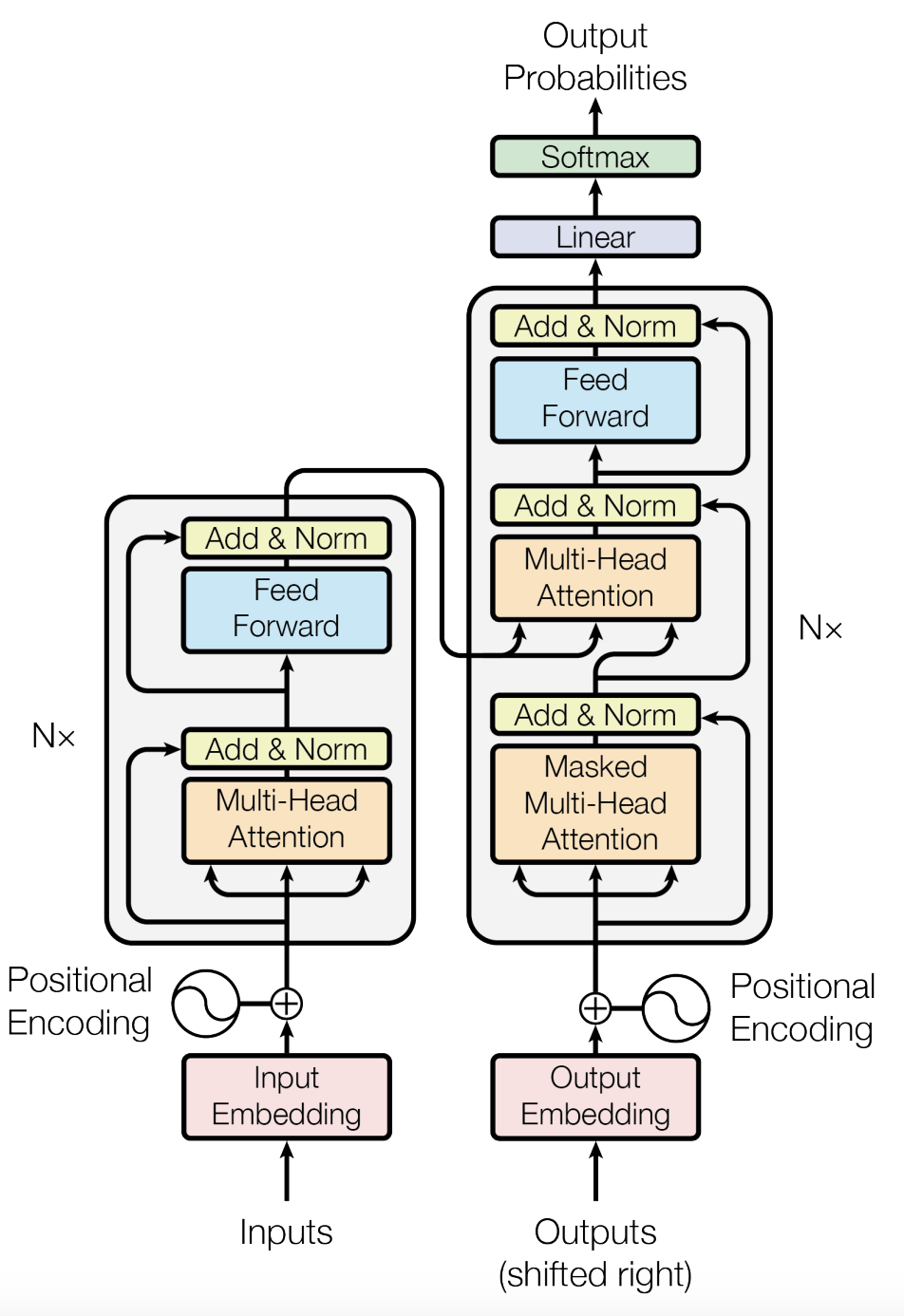}
    \caption{Transformer model architecture. The encoder is composed of six identical layers; in each layer, there are two sublayers, one of which is the multi-head self-attention mechanism \cite{attention}.}
\end{figure}

Various transformer models have been developed, typically intended to increase performance on GLUE benchmark tests, which tests natural language understanding based on ten different tasks, such as sentiment analysis and question answering \cite{glue}. Additionally, while some transformers like GPT-3 are trained on large datasets like Common Crawl and with hundreds of billions of parameters \cite{gpt-3}, others like DistilBERT are designed on the smaller scale and are easier to implement in practical contexts \cite{Distilbert}. 

\subsection{BERT}
BERT (Bidirectional Encoder Representations from Transformers) is a transformer language model which uses transfer learning to learn unsupervised from a corpus on unlabeled data; the size of the dataset allows BERT to overcome common model weaknesses like overfitting and underfitting. When BERT is fine-tuned with a smaller set of labeled data, it can be used for a variety of natural language tasks like question answering and sentiment analysis. BERT's architecture is largely similar across different tasks, making it an effective model for less common language tasks like intent recognition. BERT has achieved novel performance on GLUE benchmark testing, a test of model performance based on ten natural language understanding tasks \cite{glue}, with a base score of 79.6 for BERT$_{BASE}$ and 82.1 for BERT$_{LARGE}$ \cite{bert}. When using Build-a-Bot, students train their own BERT model for intent recognition based on questions that they write and label.

\subsection{DistilBERT}
\begin{figure}[!h]
    \centering
    \includegraphics[scale=0.5]{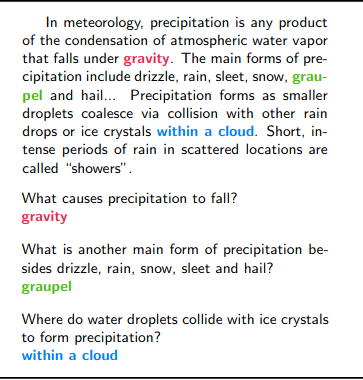}
    \caption{Question-and-answer pairs with corresponding SQuAD context \cite{squad}. Note that most answers are a few words at most -- students are encouraged to compare results with the generative T5 model and learn to use policy to make the resulting answers more conversation-like.}
    \label{squad}
\end{figure}

DistilBERT is a model with the same general architecture as BERT, but that uses distillation during pretraining in order to reduce the number of layers and make the model more lightweight. DistilBERT reduces BERT's size by forty percent while increasing speed by sixty percent and retaining ninety-seven percent of BERT's language understanding capacities, making it an effective transformer for practical applications, like training with fewer computational resources \cite{Distilbert}. DistilBERT is used for the extractive question answering step in Build-a-Bot. Students use a DistilBERT model pretrained on the Stanford Question Answering Dataset, seen in Figure \ref{squad}, a dataset consisting of over one hundred thousands questions about Wikipedia articles where the answer is a segment of text from the corresponding article \cite{squad}, rather than training the model themselves.

\subsection{T5}
\begin{figure}[!h]
    \centering
    \includegraphics[width=\linewidth]{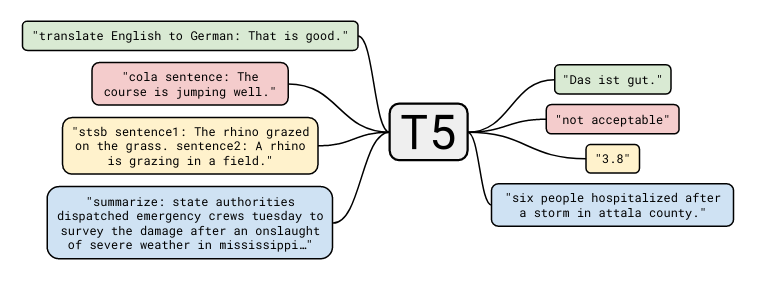}
    \caption{Text-to-text framework. All tasks are converted to an input text to output text format, including tasks like machine translation, summarization, and sentiment analysis \cite{t5}.}
    \label{t5}
\end{figure}

T5 (Text-To-Text Transfer Transformer) is a transformer model that converts all downstream language tasks, such as question answering, sentence completion, and sentiment analysis, to a text-to-text format (demonstrated in Figure \ref{t5}), in contrast with BERT models that can only output a span of the input or a class label. T5 achieved a state-of-the-art GLUE score of 85.97 in 2019 \cite{t5}. A pretrained T5 model is used for the generative question answering step in Build-a-Bot.

\section{Intent Recognition and Question Answering Model Architecture}

\subsection{Pipeline Overview}
The language modeling pipeline is used to generate answers to student questions and consists of data labeling and collection, data augmentation, filtering, intent recognition, and question answering. The purpose of applying the pipeline design is to develop a sustainable architecture for student question answering and interaction that allows easy retraining as new data is added as students add more labeled questions to their dataset; pipeline designs are popular in the NLP community due to their ease of scalability, modifiability, and complexity \cite{pipeline}. Moreover, this design allows students to dissect and modify a system of multiple transformers and data processes, providing both a comprehensive overview of the supervised learning process and ample opportunity for students to actively tinker and explore.

\begin{figure}[!h]
    \centering
    \includegraphics[width=\linewidth]{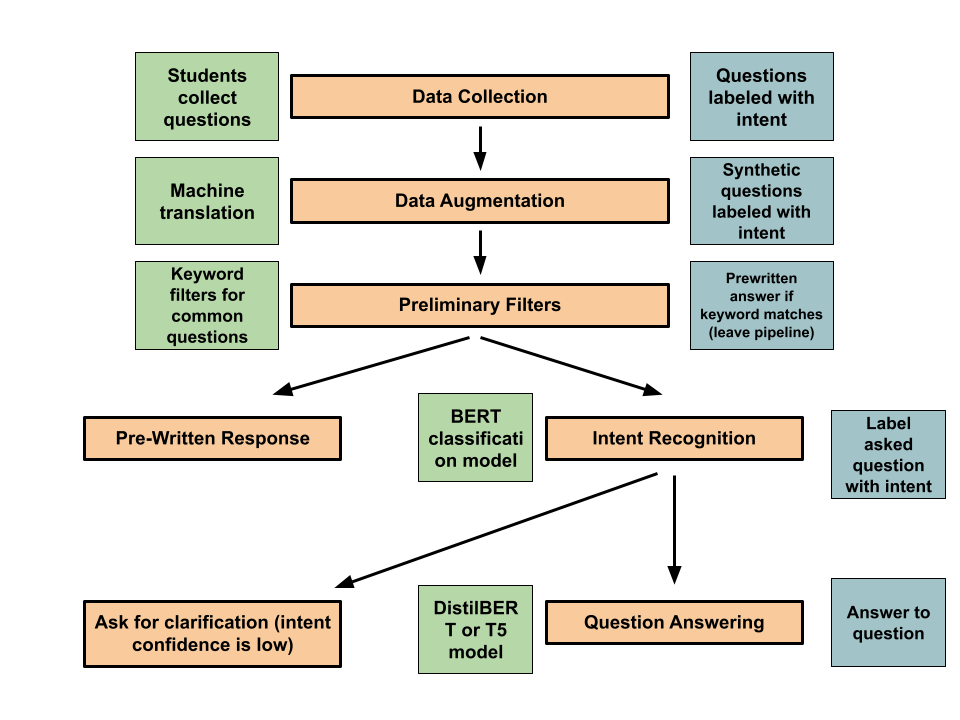}
    \caption{Diagram of the pipeline. The orange boxes show the five steps of the process, the green boxes show the methodology for each step, and the blue boxes show the output for each step. Data collection and augmentation are done before model training, while a student question will go through the three remaining steps.}
\end{figure}

\subsection{Data Collection and Labeling}
Data is collected by the students in the form of questions; these questions are then labeled by students with an intent previously set by their instructor. For example, an instructor might set six intents based on each chapter from a textbook, and, as students asked questions about the textbook material, they would upload these questions with a label corresponding to the chapter of the textbook where the answer would be found.

\subsection{Data Augmentation}
Data augmentation is the practice of creating synthetic data to add samples to a limited dataset \cite{dataAug}. It is necessary in this context due to the small size of the natural language dataset; students would find it difficult to collect hundreds of questions for each intent. Backtranslation is used in order to generate synthetic data; machine translation is used to translate questions into another language and then back to the original language, creating utterances with similar semantic meaning but different syntactic structures \cite{dataAug}. The NLPAug library is used to implement this technique on the student-created question dataset; the BacktranslationAug method is used in order to seamlessly leverage two translation models to create synthetic data \cite{nlpaug}. Moreover, the inclusion of this functionality pushes students to consider the importance of the size of their training datasets, and a high-level explanation of the technique further engages learners. 

\begin{figure}[!h]
    \centering
    \includegraphics[width=\linewidth]{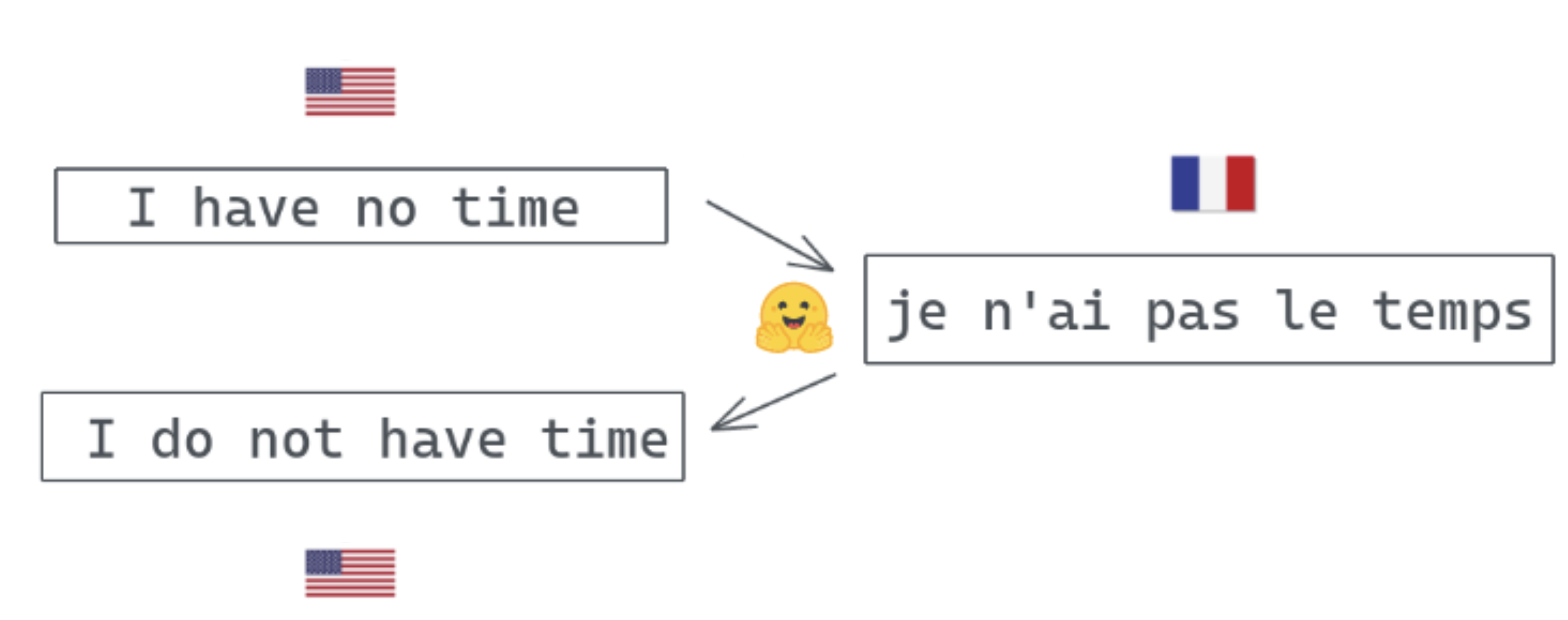}
    \caption{General example of machine translation used to create synthetic data; in this case, sample sequences are translated from English, to French, then back to English. \cite{back}}
\end{figure}

\subsection{Preliminary Filters}
Keyword filters can be added as a policy to precisely answer general technical questions, sway from distractors, and improve user experience; for example, for a class requiring students to enter a certain username and password to use digital course materials, a prewritten response specifying the format of the login could be given to any question containing the keyword "login", or if a student asks for the location of the classroom, a prewritten response giving the classroom number could be given. Questions that contain keywords do not enter the intent recognition and question answering parts of the pipeline. This step is also used to help students distinguished between AI and rule-based algorithms, as well as the importance of policy.

\subsection{Intent Recognition}
Intent recognition is used to classify which context should be used to answer a question; this helps the model give more relevant answers by using more relevant contexts. Labels for contexts are set by the instructor at the beginning of the session -- use cases might include a history class teacher specifying each intent as a different historical figure, and using a short biography of the corresponding figure as each intent's context, or a biology teacher setting each intent as a chapter of the textbook and using the text from that chapter as the context. A BERT-based architecture trained on the student question dataset is used for the intent recognition model, with the number of epochs set by the students while they learn about model training.

\subsection{Question Answering}
The final phase of the pipeline is used to answer student questions using a context based on the previously-recognized intent. Two transformer models were tested for this task: DistilBERT and T5. These models are extractive and generative, respectively; while DistilBERT answers are exclusively comprised of spans of tokens from the context, T5 answers are generated based on the given context. DistilBERT's lightweight nature makes it ideal for practical application in the classroom context; answer generation is speedy and not computationally taxing. Unlike the intent recognition model, which is customized to the students' questions, the question answering models are general models pretrained on the Stanford Question Answering dataset, so students only use, rather than train, these models. Students are encouraged to try training with both transformers and compare results.

\section{Build-a-Bot Tool Interface}
Build-a-Bot will be deployed as an open-source tool available on GitHub; a PySimpleGUI is used to present Python code from Jupyter notebooks accompanied by visuals, explanations, and interaction opportunities. The use of this interface allows the program to be downloaded as an executable file, making setup easy even for instructors with little technical experience. The first notebook contains introductory notes and steps for the instructor to define intents and contexts alongside students; afterwards, each step of the pipeline has its own devoted notebook for students to input data before moving onto the next step. In order to keep the tool middle-school appropriate, students are not required to modify code; rather, the main tasks revolve around uploading data and other inputs through a separate PySimpleGUI interface. The students' final chatbots are deployed in a new window using the chatbotAI library, which inputs model responses into a texting-style interface. Each notebook is also accompanied by slides that heavily incorporate images to keep student interest.

\begin{figure}[!h]
    \centering
    \includegraphics[scale=0.5]{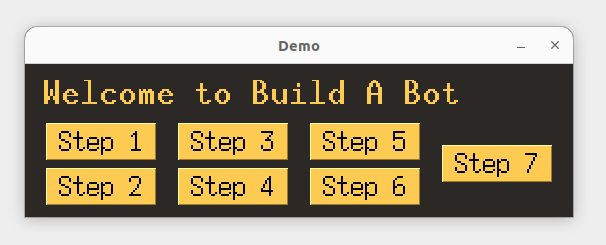}
    \caption{Demo of the interface that serves as the students' home page; they return here after completing each step to access the next one. Documentation is downloaded alongside the executable to provide information about Build-a-Bot and how to use it. Step 1 is data collection, Step 2 is data augmentation, Step 3 is policy filtering, Step 4 is intent recognition, Step 5 is extractive question answering, Step 6 is generative question answering, and Step 7 is final deployment of the models as a chatbot in a new window.}
\end{figure}

\begin{figure}[!h]
    \centering
    \includegraphics[scale=0.15]{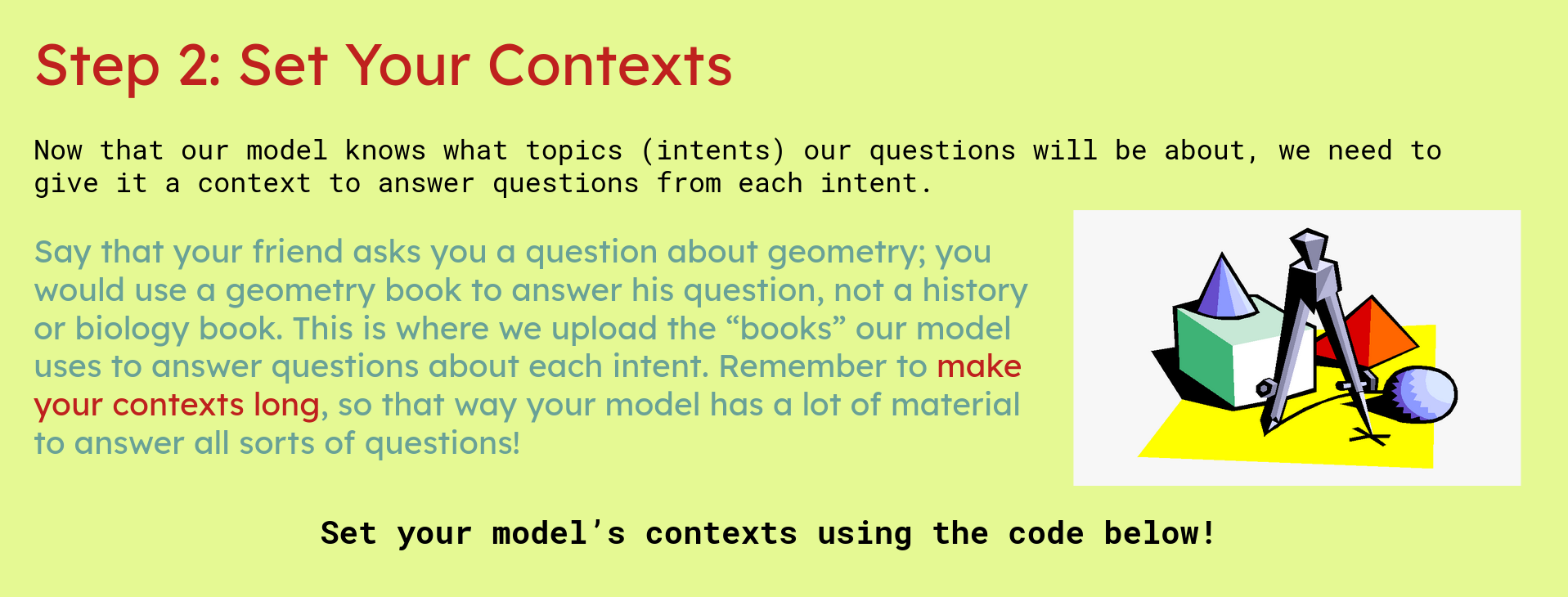}
    \caption{Example of instruction from the augmentation step. Metaphors are used to make the material suitable for a middle school audience.}
\end{figure}

\begin{figure}[!h]
    \centering
    \includegraphics[scale=0.5]{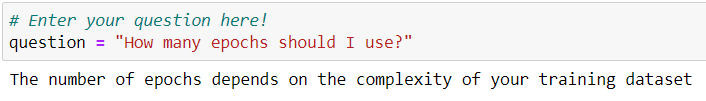}
    \caption{A student interaction from the testing phase in the generative question answering step. The chatbot's topic was supervised learning, and the recognizing intent, in this case, was model training.}
    \label{epochs}
\end{figure}

\begin{figure}[!h]
    \centering
    \includegraphics[scale=0.35]{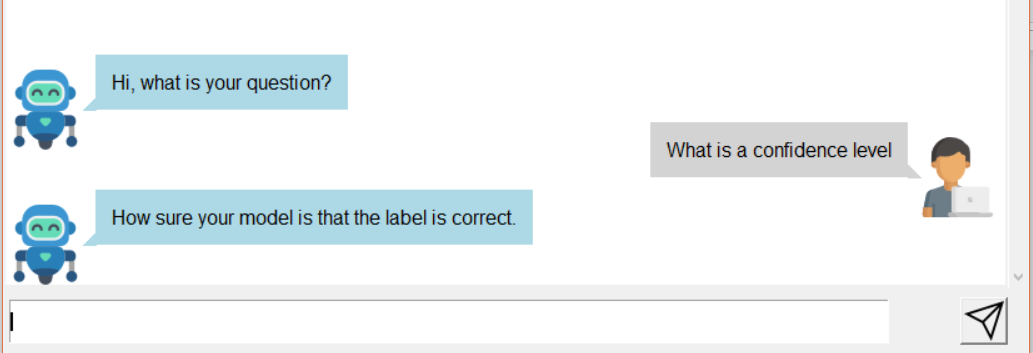}
    \caption{A student interaction with a final deployed chatbot, where the recognized intent is "model testing" and the generative T5 question answering model is used to generate an answer from a context about model testing. The chatbot is deployed in a new window using the chatbotAI library.}
\end{figure}

\begin{figure}[!h]
    \centering
    \includegraphics[scale=0.5]{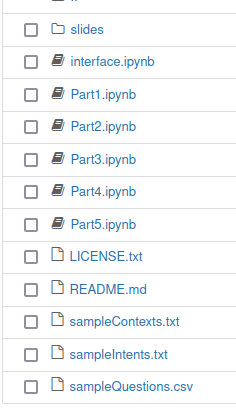}
    \caption{File structure of the tool; the interface section contains code to make all of the notebooks accessible through a PySimpleGUI interface, which is downloaded by instructors as an executable file.}
    \label{filestructure}
\end{figure}

\section{Testing Suite}
Two testing suites have been developed to use with Build-a-Bot; the first is a machine learning suite with contexts about each step of the machine learning process (e.g. data labeling, training). This suite was used in the interface images above. The second is based off of the Utah State Board of Education's open-source fifth grade science textbook \cite{textbook}, with seventy-five questions labeled with one of five intents -- Patterns in Earth's Features, Earth's Water, Weathering and Erosion, Interactions between Systems, and Impact on Humans -- which are based on earth science chapters on the textbooks. Contexts for each intent are also provided as the corresponding chapters of the textbook. This suite is provided as a sample for students and teachers to see what a good dataset for the tool looks like, and the model can also be used with the questions, contexts, and intents provided to use the tool without making an entire dataset from scratch.

\begin{figure*}[!h]
    \centering
    \includegraphics[width=\linewidth]{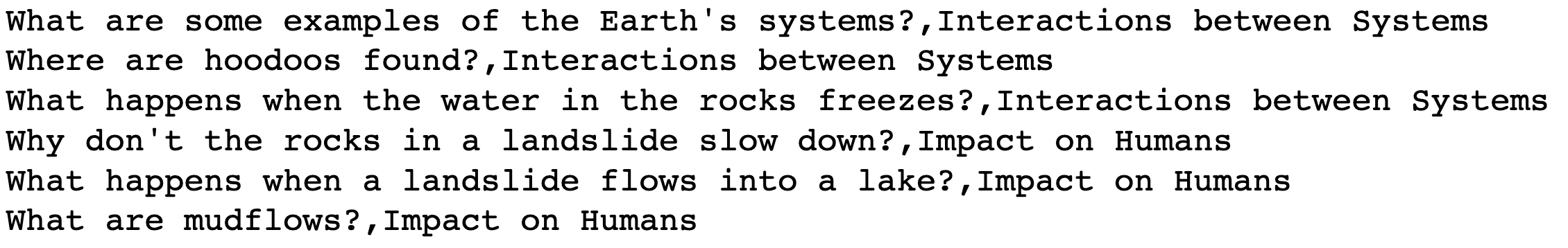}
    \caption{Some of the labeled questions given as part of the earth science testing suite. These questions can be found in the sampleQuestions.csv file seen in Figure \ref{filestructure}. The sample intent file (which lists each intent a question can be labeled as) and context file (corresponding textbook passages for each intent) are also shown as \textit{sampleIntents.txt} and \textit{sampleContexts.txt}, respectively.}
\end{figure*}

\section{Conclusion}
In this paper, we created a pipeline with multiple transformers to generate natural language answers to student questions; moreover, we used this pipeline to make an open-source tool to enable students to learn the supervised learning process by training the models in the pipeline. The tool is inexpensive to use and practical for a classroom environment, as instructors can download the tool as an executable file and deploy it without any programming knowledge. This enables educators of any level of technical experience to build their students' AI literacy with a curriculum based around constructivist learning theory. Students also build system engineering and analysis skills by engaging with and modifying the multi-step language pipeline. Further work involves presenting results of the efficacy of the language modeling used and piloting the tool in classroom settings. We also hope to expand access by making our tool usable by all operating systems and improving documentation.

\bibliography{aaai23}

\begin{thebibliography}{28}
\providecommand{\natexlab}[1]{#1}

\bibitem[{Brown et~al.(2020)Brown, Mann, Ryder, Subbiah, Kaplan, Dhariwal,
  Neelakantan, Shyam, Sastry, Askell, Agarwal, Herbert-Voss, Krueger, Henighan,
  Child, Ramesh, Ziegler, Wu, Winter, Hesse, Chen, Sigler, Litwin, Gray, Chess,
  Clark, Berner, McCandlish, Radford, Sutskever, and Amodei}]{gpt-3}
Brown, T.; Mann, B.; Ryder, N.; Subbiah, M.; Kaplan, J.~D.; Dhariwal, P.;
  Neelakantan, A.; Shyam, P.; Sastry, G.; Askell, A.; Agarwal, S.;
  Herbert-Voss, A.; Krueger, G.; Henighan, T.; Child, R.; Ramesh, A.; Ziegler,
  D.; Wu, J.; Winter, C.; Hesse, C.; Chen, M.; Sigler, E.; Litwin, M.; Gray,
  S.; Chess, B.; Clark, J.; Berner, C.; McCandlish, S.; Radford, A.; Sutskever,
  I.; and Amodei, D. 2020.
\newblock Language Models are Few-Shot Learners.
\newblock In Larochelle, H.; Ranzato, M.; Hadsell, R.; Balcan, M.; and Lin, H.,
  eds., \emph{Advances in Neural Information Processing Systems}, volume~33,
  1877--1901. Curran Associates, Inc.

\bibitem[{Devlin et~al.(2018)Devlin, Chang, Lee, and Toutanova}]{bert}
Devlin, J.; Chang, M.; Lee, K.; and Toutanova, K. 2018.
\newblock {BERT:} Pre-training of Deep Bidirectional Transformers for Language
  Understanding.
\newblock \emph{CoRR}, abs/1810.04805.

\bibitem[{Feng et~al.(2021)Feng, Gangal, Wei, Chandar, Vosoughi, Mitamura, and
  Hovy}]{dataAug}
Feng, S.~Y.; Gangal, V.; Wei, J.; Chandar, S.; Vosoughi, S.; Mitamura, T.; and
  Hovy, E.~H. 2021.
\newblock A Survey of Data Augmentation Approaches for {NLP}.
\newblock \emph{CoRR}, abs/2105.03075.

\bibitem[{Foster and Stagl(2018)}]{threeTheories}
Foster, G.; and Stagl, S. 2018.
\newblock Design, implementation, and evaluation of an inverted (flipped)
  classroom model economics for sustainable education course.
\newblock \emph{Journal of Cleaner Production}, 183: 1323--1336.

\bibitem[{Fry, Kennedy, and Funk(2021)}]{stem}
Fry, R.; Kennedy, B.; and Funk, C. 2021.
\newblock STEM jobs see uneven progress in increasing gender, racial and ethnic
  diversity.
\newblock \emph{Pew Research Center}.

\bibitem[{Fu et~al.(2022)Fu, Gao, Zhao, rong Wen, and Yan}]{convoAI}
Fu, T.; Gao, S.; Zhao, X.; rong Wen, J.; and Yan, R. 2022.
\newblock Learning towards conversational AI: A survey.
\newblock \emph{AI Open}, 3: 14--28.

\bibitem[{Gorton et~al.(2011)Gorton, Wynne, Liu, and Yin}]{pipeline}
Gorton, I.; Wynne, A.; Liu, Y.; and Yin, J. 2011.
\newblock Components in the Pipeline.
\newblock \emph{IEEE Software}, 28(3): 34--40.

\bibitem[{J.~Bancifra(2022)}]{studentCenter}
J.~Bancifra, J. 2022.
\newblock Supervisory practices of Department Heads and teachers' performance:
  Towards A proposed Enhancement Program.
\newblock \emph{APJAET - Journal ay Asia Pacific Journal of Advanced Education
  and Technology}, 25--33.

\bibitem[{Kim et~al.(2021)Kim, Jang, Kim, Choi, Jung, Kim, and
  Kim}]{competencies}
Kim, S.; Jang, Y.; Kim, W.; Choi, S.; Jung, H.; Kim, S.; and Kim, H. 2021.
\newblock Why and What to Teach: AI Curriculum for Elementary School.
\newblock \emph{Proceedings of the AAAI Conference on Artificial Intelligence},
  35(17): 15569--15576.

\bibitem[{Kiperwasser and Goldberg(2016)}]{bidirectionality}
Kiperwasser, E.; and Goldberg, Y. 2016.
\newblock {Simple and Accurate Dependency Parsing Using Bidirectional LSTM
  Feature Representations}.
\newblock \emph{Transactions of the Association for Computational Linguistics},
  4: 313--327.

\bibitem[{Li, Lund, and Nordsteien(2021)}]{flippedActive}
Li, R.; Lund, A.; and Nordsteien, A. 2021.
\newblock The link between flipped and active learning: a scoping review.
\newblock \emph{Teaching in Higher Education}, 0(0): 1--35.

\bibitem[{Ma(2019)}]{nlpaug}
Ma, E. 2019.
\newblock NLPAug Python Library.
\newblock \url{https://github.com/makcedward/nlpaug}.

\bibitem[{Piaget(1955)}]{piaget}
Piaget, J. 1955.
\newblock \emph{The Construction of Reality in the Child}.
\newblock Routledge.

\bibitem[{Prince(2004)}]{prince}
Prince, M. 2004.
\newblock Does active learning work? A review of the research.
\newblock \emph{J. Eng. Educ.}, 93(3): 223--231.

\bibitem[{Raffel et~al.(2022)Raffel, Shazeer, Roberts, Lee, Narang, Matena,
  Zhou, Li, and Liu}]{t5}
Raffel, C.; Shazeer, N.; Roberts, A.; Lee, K.; Narang, S.; Matena, M.; Zhou,
  Y.; Li, W.; and Liu, P.~J. 2022.
\newblock Exploring the Limits of Transfer Learning with a Unified Text-to-Text
  Transformer.
\newblock \emph{J. Mach. Learn. Res.}, 21(1).

\bibitem[{Rajpurkar et~al.(2016)Rajpurkar, Zhang, Lopyrev, and Liang}]{squad}
Rajpurkar, P.; Zhang, J.; Lopyrev, K.; and Liang, P. 2016.
\newblock SQuAD: 100, 000+ Questions for Machine Comprehension of Text.
\newblock \emph{CoRR}, abs/1606.05250.

\bibitem[{Sabuncuoglu(2020)}]{middleCurriculum}
Sabuncuoglu, A. 2020.
\newblock Designing One Year Curriculum to Teach Artificial Intelligence for
  Middle School.
\newblock In \emph{Proceedings of the 2020 ACM Conference on Innovation and
  Technology in Computer Science Education}, ITiCSE '20, 96–102. New York,
  NY, USA: Association for Computing Machinery.
\newblock ISBN 9781450368742.

\bibitem[{Sanh et~al.(2019)Sanh, Debut, Chaumond, and Wolf}]{Distilbert}
Sanh, V.; Debut, L.; Chaumond, J.; and Wolf, T. 2019.
\newblock DistilBERT, a distilled version of {BERT:} smaller, faster, cheaper
  and lighter.
\newblock \emph{CoRR}, abs/1910.01108.

\bibitem[{Sennrich, Haddow, and Birch(2016)}]{back}
Sennrich, R.; Haddow, B.; and Birch, A. 2016.
\newblock Improving Neural Machine Translation Models with Monolingual Data.
\newblock In \emph{Proceedings of the 54th Annual Meeting of the Association
  for Computational Linguistics (Volume 1: Long Papers)}, 86--96. Berlin,
  Germany: Association for Computational Linguistics.

\bibitem[{Slavich and Zimbardo(2012)}]{transform}
Slavich, G.~M.; and Zimbardo, P.~G. 2012.
\newblock Transformational teaching: Theoretical underpinnings, basic
  principles, and core methods.
\newblock \emph{Educ. Psychol. Rev.}, 24(4): 569--608.

\bibitem[{Tao et~al.(2021)Tao, Feng, Yan, Wu, and Jiang}]{convoAI2}
Tao, C.; Feng, J.; Yan, R.; Wu, W.; and Jiang, D. 2021.
\newblock A Survey on Response Selection for Retrieval-based Dialogues.
\newblock In Zhou, Z.-H., ed., \emph{Proceedings of the Thirtieth International
  Joint Conference on Artificial Intelligence, {IJCAI-21}}, 4619--4626.
  International Joint Conferences on Artificial Intelligence Organization.
\newblock Survey Track.

\bibitem[{Touretzky et~al.(2019{\natexlab{a}})Touretzky, Gardner-McCune,
  Breazeal, Martin, and Seehorn}]{bigideasgraphic}
Touretzky, D.; Gardner-McCune, C.; Breazeal, C.; Martin, F.; and Seehorn, D.
  2019{\natexlab{a}}.
\newblock A Year in K-12 {AI} Education.
\newblock \emph{{AI} Magazine}, 40(4): 88--90.

\bibitem[{Touretzky et~al.(2019{\natexlab{b}})Touretzky, Gardner-Mccune,
  Martin, and Seehorn}]{bigideas}
Touretzky, D.~S.; Gardner-Mccune, C.; Martin, F.~G.; and Seehorn, D.~W.
  2019{\natexlab{b}}.
\newblock Envisioning AI for K-12: What Should Every Child Know about AI?
\newblock In \emph{AAAI Conference on Artificial Intelligence}.

\bibitem[{{Utah State Board of Education OER}(2022)}]{textbook}
{Utah State Board of Education OER}. 2022.
\newblock \emph{5th Grade for Utah SEEd Standards}.
\newblock CK-12 Foundation.

\bibitem[{Vaswani et~al.(2017)Vaswani, Shazeer, Parmar, Uszkoreit, Jones,
  Gomez, Kaiser, and Polosukhin}]{attention}
Vaswani, A.; Shazeer, N.; Parmar, N.; Uszkoreit, J.; Jones, L.; Gomez, A.~N.;
  Kaiser, L.; and Polosukhin, I. 2017.
\newblock Attention Is All You Need.
\newblock \emph{CoRR}, abs/1706.03762.

\bibitem[{Wang et~al.(2018)Wang, Singh, Michael, Hill, Levy, and Bowman}]{glue}
Wang, A.; Singh, A.; Michael, J.; Hill, F.; Levy, O.; and Bowman, S. 2018.
\newblock {GLUE}: A Multi-Task Benchmark and Analysis Platform for Natural
  Language Understanding.
\newblock In \emph{Proceedings of the 2018 {EMNLP} Workshop {B}lackbox{NLP}:
  Analyzing and Interpreting Neural Networks for {NLP}}, 353--355. Brussels,
  Belgium: Association for Computational Linguistics.

\bibitem[{Williams et~al.(2022)Williams, Ali, Devasia, DiPaola, Hong, Kaputsos,
  Jordan, and Breazeal}]{ethics}
Williams, R.; Ali, S.; Devasia, N.; DiPaola, D.; Hong, J.; Kaputsos, S.~P.;
  Jordan, B.; and Breazeal, C. 2022.
\newblock {AI} + ethics curricula for middle school youth: Lessons learned from
  three project-based curricula.
\newblock \emph{Int. J. Artif. Intell. Educ.}, 1--59.

\bibitem[{Wollowski et~al.(2016)Wollowski, Selkowitz, Brown, Goel, Luger,
  Marshall, Neel, Neller, and Norvig}]{aiEdSurvey}
Wollowski, M.; Selkowitz, R.; Brown, L.; Goel, A.; Luger, G.; Marshall, J.;
  Neel, A.; Neller, T.; and Norvig, P. 2016.
\newblock \emph{Proceedings of the AAAI Conference on Artificial Intelligence},
  30(1).

\end{thebibliography}

\end{document}